\documentclass[runningheads]{llncs}

\usepackage{eccv}

\usepackage{eccvabbrv}

\usepackage{graphicx}
\usepackage{booktabs}

\usepackage[accsupp]{axessibility}  

\usepackage{color}
\usepackage[normalem]{ulem}
\usepackage{rotating}
\usepackage{tabularray}

\usepackage{makecell}
\usepackage{array}
\usepackage{multirow}
\usepackage{ragged2e}

\usepackage{multirow}

\usepackage{hyperref}

\usepackage{orcidlink}

\usepackage[nolist,nohyperlinks]{acronym}

\acrodef{2d}[2D]{two-dimensional}
\acrodef{3d}[3D]{three-dimensional}
\acrodef{sfm}[SfM]{structure-from-motion}
\acrodef{sift}[SIFT]{Scale-Invariant Feature Transform}
\acrodef{ransac}[RANSAC]{random sample consensus}
\acrodef{lift}[LIFT]{Learned Invariant Feature Transform}
\acrodef{cnn}[CNN]{convolutional neural network}
\acrodef{dl}[DL]{deep learning}
\acrodef{fast}[FAST]{Features from Accelerated Segment Test}
\acrodef{surf}[SURF]{Speeded-Up Robust Features}
\acrodef{dog}[DoG]{Difference-of-Gaussians}
\acrodef{orb}[ORB]{Oriented FAST and Rotated BRIEF}
\acrodef{aliked}[ALIKED]{A LIghter Keypoint and Descriptor Extraction with Deformable transformation}
\acrodef{nms}[NMS]{non-maximum suppression}
\acrodef{dkd}[DKD]{differentiable keypoint detection}
\acrodef{dedode}[DeDoDe]{Detect, Don't Describe -- Describe, Don't Detect for Local Feature Matching}
\acrodef{nn}[NN]{Nearest Neighbour}
\acrodef{mnn}[MNN]{Mutual Nearest Neighbour}
\acrodef{dsm}[DSM]{Dual Softmax Matcher}
\acrodef{gnn}[GNN]{graph neural network}
\acrodef{roma}[RoMa]{Robust Dense Feature Matching}
\acrodef{imc24}[IMC24]{Image Matching Challenge 2024}
\acrodef{dexined}[DexiNed]{Dense Extreme Inception Network for Edge Detection}
\acrodef{sota}[SOTA]{state-of-the-art}
\acrodef{auc}[AUC]{Area Under the Curve}
\acrodef{brief}[BRIEF]{Binary Robust Independent Elementary Features}

\makeatletter
\newcommand{\printfnsymbol}[1]{%
  \textsuperscript{\@fnsymbol{#1}}%
}
\makeatother

\begin{document}

\title{Mismatched: Evaluating the Limits of Image Matching Approaches and Benchmarks} 

\titlerunning{Mismatched: Limits of Image Matching Approaches and Benchmarks}

\author{Sierra~Bonilla\thanks{These authors contributed equally to this work and are co-first authors}\orcidlink{0009-0008-6384-811X} \and
Chiara~Di~Vece\printfnsymbol{1}\orcidlink{0000-0002-2853-9791} \and
Rema~Daher\printfnsymbol{1}\orcidlink{0000-0002-0740-7490} \and
Xinwei~Ju\printfnsymbol{1}\orcidlink{0000-0002-1137-9357} \and  Danail~Stoyanov\orcidlink{0000-0002-0980-3227} \and Francisco~Vasconcelos\orcidlink{0000-0002-4609-1177} \and Sophia~Bano\orcidlink{0000-0003-1329-4565}}

\authorrunning{S.~Bonilla, C.~Di~Vece, R.~Daher, X.~Ju et al.}

\institute{Department of Computer Science and UCL Hawkes Institute,\\University College London, London WC1E 6BT, UK\\
\email{\{sierra.bonilla.21,chiara.divece.20,rema.daher.20,xinwei.ju.22\}@ucl.ac.uk}}

\maketitle

\begin{abstract}
  Three-dimensional (3D) reconstruction from two-dimensional images is an active research field in computer vision, with applications ranging from navigation and object tracking to segmentation and three-dimensional modeling. Traditionally, parametric techniques have been employed for this task. However, recent advancements have seen a shift towards learning-based methods. Given the rapid pace of research and the frequent introduction of new image matching methods, it is essential to evaluate them. In this paper, we present a comprehensive evaluation of various image matching methods using a structure-from-motion pipeline. We assess the performance of these methods on both in-domain and out-of-domain datasets, identifying key limitations in both the methods and benchmarks. We also investigate the impact of edge detection as a pre-processing step. Our analysis reveals that image matching for 3D reconstruction remains an open challenge, necessitating careful selection and tuning of models for specific scenarios, while also highlighting mismatches in how metrics currently represent method performance. Code is available at \href{https://github.com/surgical-vision/colmap-match-converter}{https://github.com/surgical-vision/colmap-match-converter}.
\end{abstract}

\section{Introduction}
\label{sec:intro}

Reconstructing \ac{3d} structures from \ac{2d} images has been a fundamental challenge in computer vision for decades. The classic \ac{sfm} technique, first proposed by Ullman in the 1970s\cite{ullman1979interpretation}, involves identifying distinct feature points, matching points with overlapping views, and solving the triangulation problem to estimate the \ac{3d} feature points in space as well as the camera locations and orientations. This process has historically been addressed through a series of smaller problems: image retrieval, feature extraction, feature matching, geometric verification, camera pose estimation and outlier rejection using geometric \ac{ransac}\cite{fischler1981random}, triangulation, and bundle adjustment. 

From \ac{sfm}, sparse clouds and paired image poses can be obtained; these are used for a large range of computer vision tasks including, but not limited to, dense \ac{3d} reconstruction\cite{furukawa2009accurate,geiger2011stereoscan}, autonomous navigation and guidance\cite{hane2011stereo,geiger2012we}, and \ac{3d} object detection, tracking and segmentation\cite{antoun2020towards,voigtlaender2019mots}. Various fields benefit from these sparse clouds and poses, such as autonomous driving\cite{geiger2012we}, archaeology and architecture\cite{bagnolo2019hbim}, manufacturing\cite{khan2021vision}, healthcare\cite{yang20243d}, and urban planning\cite{iheaturu2022simplified}. 

A key component of \ac{sfm} is the feature extraction and matching step, which aims to identify corresponding points across multiple images. This is a challenging task, even for humans, as it requires reliably identifying the same geometric points across multiple images, especially under conditions such as varying camera parameters, illumination, seasonal changes, occlusions, and transparency. In the classic \ac{sfm} technique, \ac{sift}, developed by Lowe in 1999\cite{lowe1999object} as a parametric method, was used for correspondence matching and remains the most predominantly used method across real-world settings\cite{zhang2023deep,borhani2019digital}. However, traditional parametric methods, like \ac{sift} and \ac{surf}\cite{bayherbert} often struggle with repeating patterns or texture-less spaces. In recent years, \ac{dl} approaches like \ac{aliked}, SuperPoint, and DISK have emerged as potential alternatives to \ac{sift} for feature extraction and matching\cite{zhao2023aliked,detone2018superpoint,tyszkiewicz2020disk}. These methods aim to learn robust and accurate features or correspondences. However, they can falter when applied to out-of-domain data, where image characteristics differ significantly from the training data, such as with transparent objects or drastic lighting changes\cite{jiang2024omniglue}. Available training datasets are not comprehensive due to the expensive and time-consuming nature of acquiring large-scale, high-quality, and diverse \ac{3d} scenes. Consequently, models overfit to the specific characteristics of the available training data and struggle to generalize to new domains. Despite these challenges, several recent works have claimed to address the generalizability issue and outperform traditional methods like \ac{sift} \cite{jiang2024omniglue,potje2024xfeat,edstedt2024roma,shen2024gim}. 

\begin{figure}[!t]
    \centering
    \includegraphics[width=0.6\linewidth]{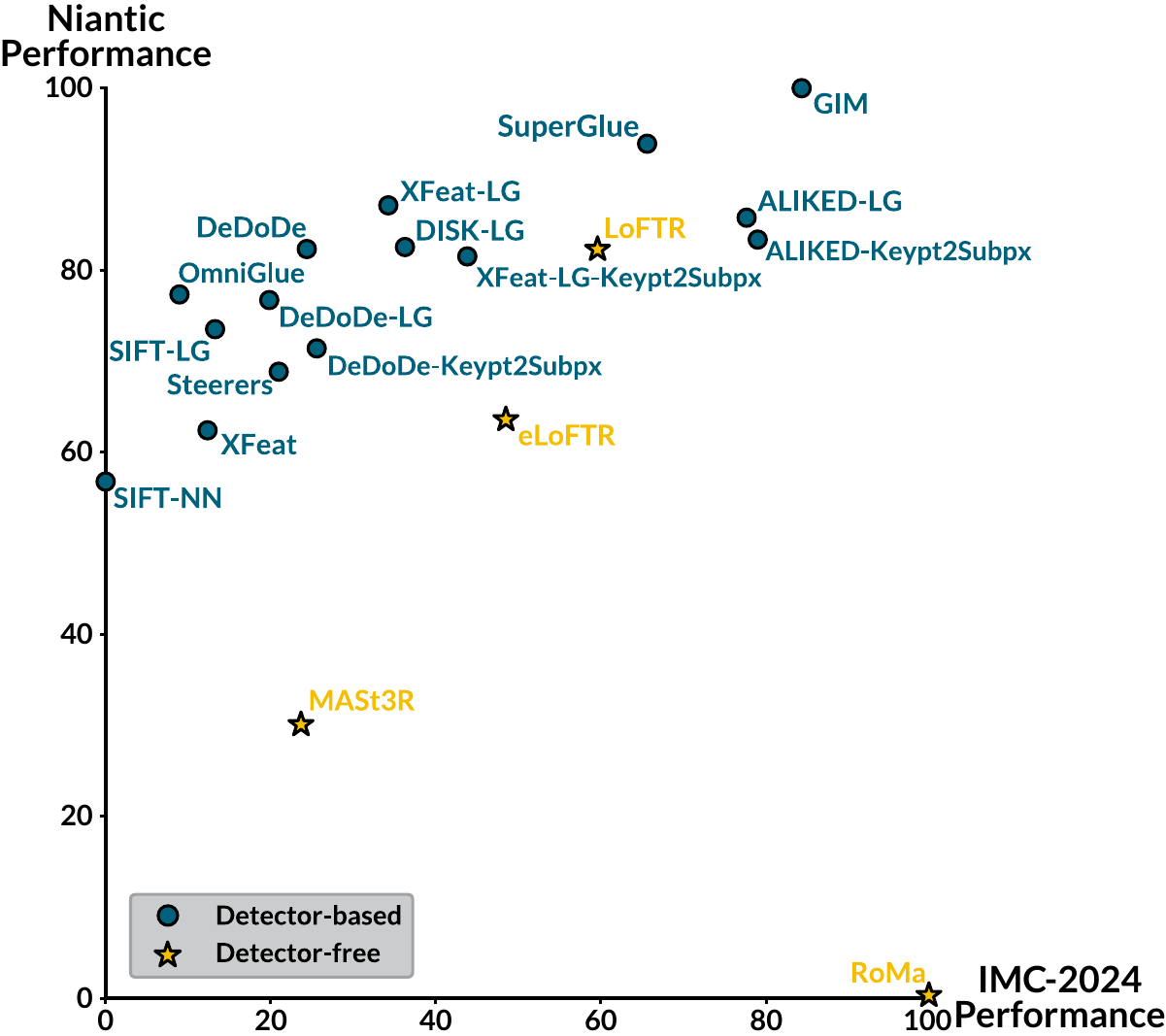}
    \caption{Normalized $mAA$ for out-of-domain (IMC24)\cite{image-matching-challenge-2024} versus in-domain (Niantic) datasets\cite{arnold2022map}.}
    \label{fig:intro}
\end{figure}

This paper aims to provide insights into the most recent feature extraction and matching methods for \ac{3d} reconstruction from \ac{2d} images. To the best of our knowledge, this paper is the first to comprehensively evaluate recent \ac{sota} models. We discuss the importance of generalizability for real-world applications and offer recommendations for research directions in this field. The contributions of our work can be summarized as:
\begin{enumerate}
    \item Comparing 20 \ac{sota} image matching methods, 8 of which were presented in 2024, and discussing their generalizability and current dataset limitations (\Cref{fig:intro}). We use (a) an in-domain dataset consisting of clean images from a single scene with consistent lighting conditions and camera parameters \cite{arnold2022map} and (b) an out-of-domain dataset containing a diverse set of images with varying lighting, camera intrinsics, seasons, and object transparency \cite{image-matching-challenge-2024}.
    \item Investigating the effects of edge detection on the various traditional and \ac{dl}-based image matching methods within an \ac{sfm} pipeline using \ac{dexined}\cite{dexined_ext2023}.
    \item Evaluating the $mAA$ metric across different scenarios and its implications for assessing image matching. We examine how unregistered images influence metric outcomes and offer recommendations for clearer and more consistent metric reporting to enhance comparability.
\end{enumerate}

\section{Related Work}

Numerous studies have investigated approaches to improve feature extraction and matching for \ac{3d} reconstruction\cite{Xu_2024}. This review focuses on two primary tasks: feature extraction and matching. Feature extraction can be further broken down into feature detection and description, though many algorithms address both simultaneously. Feature detection involves locating key points in an image, identified by their $(u,v)$ coordinates, using various criteria such as edges or corners. Feature description encodes the local image region around each detected key point into a numerical vector, or descriptor, capturing information about the neighborhood. These descriptors must be invariant to properties not inherent to an object in the scene as the camera moves from one frame to another. Such properties include scale, rotation, and illumination changes. The matching task involves finding correspondences between features across multiple images by comparing descriptors. This process aims to accurately identify pairs of corresponding features representing the same point in the \ac{3d} scene. Some \ac{dl} methods attempt to address both tasks end-to-end and are commonly referred to as detector-free methods. It is worth noting that there are other approaches to estimating camera pose directly\cite{barrosolaguna2024matching2dimages3d,xiang2017posecnn}; however, we do not evaluate against such methods as we focus on feature extraction and matching.

\subsection{Feature Extraction}

The original traditional method for feature extraction, \ac{sift}, detects scale-space extrema by convolving the image with Gaussian filters at different scales and identifying keypoints as maxima or minima in the \ac{dog} map. Descriptors are computed based on gradient orientations in the local neighborhood. Although \ac{sift} is still the most well-known handcrafted feature extraction method, it is computationally intensive, making it less suitable for real-time applications or devices with limited resources. Thus, alternatives have been presented such as \ac{fast}\cite{rosten2006machine} and \ac{surf}\cite{bayherbert} (ECCV `06). \ac{fast} faces limitations in settings with significant scale changes and \ac{surf} struggles with extreme rotations because of approximations used in its keypoint detection and orientation assignment. As a solution, \Ac{orb}\cite{rublee2011orb}, introduced in 2011, combines the \ac{fast} keypoint detector and the \ac{brief} feature descriptor\cite{calonder2010brief} which attempts to address the computational inefficiencies of previous methods as well as the lack of orientation and scale invariance. However, as with other traditional methods, \ac{orb} tends to underperform in low-textured regions, repeated patterns, occlusions, and significant illumination changes.

Seeking to overcome the limitations of handcrafted feature extractors by leveraging \ac{dl} techniques, \ac{lift} (ECCV `16)\cite{yi2016lift}, one of the first popular learning-based methods, integrates keypoint detection, orientation estimation, and feature description into a single optimizable framework. \ac{lift} combines three \acp{cnn} allowing for a fully supervised end-to-end training. However, the need for extensive handcrafted labeled data limits this approach. Instead of relying solely on supervised learning, DISK\cite{tyszkiewicz2020disk} uses a reward-based system to refine keypoints sampled from \ac{cnn}-generated heatmaps, training correct matches based on geometric ground-truth with a lower need for handcrafted labeled data. In contrast to full or weak supervision, SuperPoint\cite{detone2018superpoint} introduced a self-supervised convolutional model trained via image pairs generated through Homographic Adaptations. 

As an alternative, \ac{aliked}\cite{zhao2023aliked}, introduced in 2023, builds on its popular predecessor, ALIKE\cite{zhao2022alikeaccuratelightweightkeypoint}, uses deformable convolutions to model geometric transformations more flexibly, and replaces \ac{nms} with \ac{dkd} to backpropagate gradients and produce keypoints at subpixel levels. 
In the same year, \ac{dedode}\cite{edstedt2023dedodedetectdont} was introduced; it took a different approach by decoupling keypoint detection from descriptor learning, employing a fully supervised learning approach with \acp{cnn} trained on large-scale \ac{sfm}. More recently, XFeat (CVPR `24)\cite{potje2024xfeat}, another \ac{cnn}-based feature extractor, proposed a lightweight architecture trained on large-scale datasets with pixel-level ground truth correspondences, progressively increasing the number of channels to reduce computational load. Both \ac{dedode} and XFeat can be used with a handcrafted matching method for end-to-end matching. 

\subsection{Feature Matching}

Traditionally, keypoints are matched between images by comparing descriptors using a \ac{nn} approach, which takes the Euclidean distance between the descriptor vectors\cite{eppstein1998raising}. \ac{mnn}\cite{juan1982programme} is an extension to \ac{nn} that considers a keypoint match valid if it is the nearest neighbor in both directions; this mutual check reduces mismatches. The \ac{dsm} is another handcrafted matching method that considers both row-wise and column-wise normalization of the matching matrix. It ensures that the matched features are mutually exclusive, where each keypoint is matched to only one keypoint in the other image, and a match is considered valid only if both keypoints are each other's best match, ensuring bidirectional consistency.

Feature matching has also been significantly impacted by the \ac{dl} revolution. Among the most notable of the methods, SuperGlue\cite{sarlin2020superglue} utilizes a \ac{gnn} to optimize the matching process by employing self-attention and cross-attention mechanisms inspired by Transformer architectures to leverage spatial relationships. As an extension of SuperGlue, LightGlue\cite{Lindenberger_2023_ICCV} enhances efficiency by introducing an adaptive matching strategy that dynamically adjusts the network size based on the difficulty of the matching problem.

OmniGlue\cite{jiang2024omniglue} and Steerers\cite{bokman2024steerers} (CVPR `24) represent the latest attempts at learning-based feature matchers. OmniGlue incorporates frozen pre-trained embeddings from DINOv2\cite{oquab2023dinov2}, a self-supervised vision transformer foundation model, to construct inter-image graphs. It also uses SuperPoint extracted keypoints to construct intra-image graphs and employs a transformer-based architecture with a novel keypoint position-guided attention mechanism to improve cross-domain generalization. Steerers proposes a solution to achieve rotation-invariant keypoint descriptions by detecting keypoints using other methods and optimizing a linear transformation that directly modifies the keypoint descriptors to encode rotations. They then employ a Max Similarity Matcher, which iteratively rotates keypoints to find maximum similarity.

\subsection{Detector-Free Matcher}

Detector-free methods bypass traditional keypoint detection and directly produce dense descriptors or feature matches across image pairs. A well-known detector-free method is LoFTR\cite{sun2021loftr}, a fully supervised transformer-based approach, that uses self-attention and cross-attention layers to generate feature descriptors conditioned on both images. However, since it is computationally intensive, Efficient LoFTR\cite{wang2024efficient} was proposed to improve efficiency through aggregated attention mechanisms and adaptive token selection. \ac{roma}\cite{edstedt2024roma} (CVPR `24) uses frozen pre-trained coarse embeddings of DINOv2 with specialized \ac{cnn} fine features. It employs a transformer-based match decoder to predict anchor probabilities instead of coordinates and introduces a loss formulation combining regression-by-classification for coarse matches and robust regression for refinement. DUSt3R\cite{wang2024dust3r} (CVPR `24) shifts from \ac{2d} to \ac{3d} space, treating matching as a \ac{3d} task using point clouds for implicit correspondence learning. This method uses a transformer-based architecture and is fully supervised. It leverages pre-trained vision transformers\cite{weinzaepfel2023crocoselfsupervisedpretraining3d} for encoding and decoding images and generating dense \ac{3d} pointmaps. MASt3R\cite{leroy2024grounding} adds a second head to the DUSt3R network to output dense local features, trained with another matching loss. However, both DUSt3R and MASt3R have high computational complexity. 

\subsection{Training Framework}

Other techniques propose training frameworks that can be utilized alongside the previously mentioned methods to improve performance. For instance, GIM\cite{shen2024gim} (ICLR `24) uses DINOv2 to build inter-image graphs acting as an end-to-end training technique when combined with other methods. GIM is not used alone but serves as an augmentative framework and claims to generalize better. Keypt2Subpx\cite{kim2024learningmakekeypointssubpixel} (ECCV `24) is designed to enhance existing methods but is also not standalone. It can be integrated into any detector to achieve sub-pixel precision by learning an offset vector for detected features and eliminate the need for specialized sub-pixel accurate detectors. The approach employs a detector-agnostic Keypoint Refinement module guided by local feature descriptors. It uses supervised training on datasets with pixel-level ground-truth correspondences.

Traditional methods for feature extraction and matching, such as \ac{sift}, are limited in efficiency and performance in regions with low textures or repeating patterns. In addition, learning-based approaches, while offering enhanced robustness and performance, require extensive labeled datasets, which limit their generalizability and come with high computational demands. Thus, despite numerous advancements, challenges in generalizability and scalability remain the central obstacles to truly replacing handcrafted methods in real-world applications. In the past year alone, over 8 new methods have been proposed to tackle them. This paper aims to cut through the noise and compare the latest models, critically evaluating whether modern techniques have truly resolved these issues.

\section{Methodology} 
\label{sec:methodology}

\begin{figure}[!t]
    \centering
    \includegraphics[width=\linewidth]{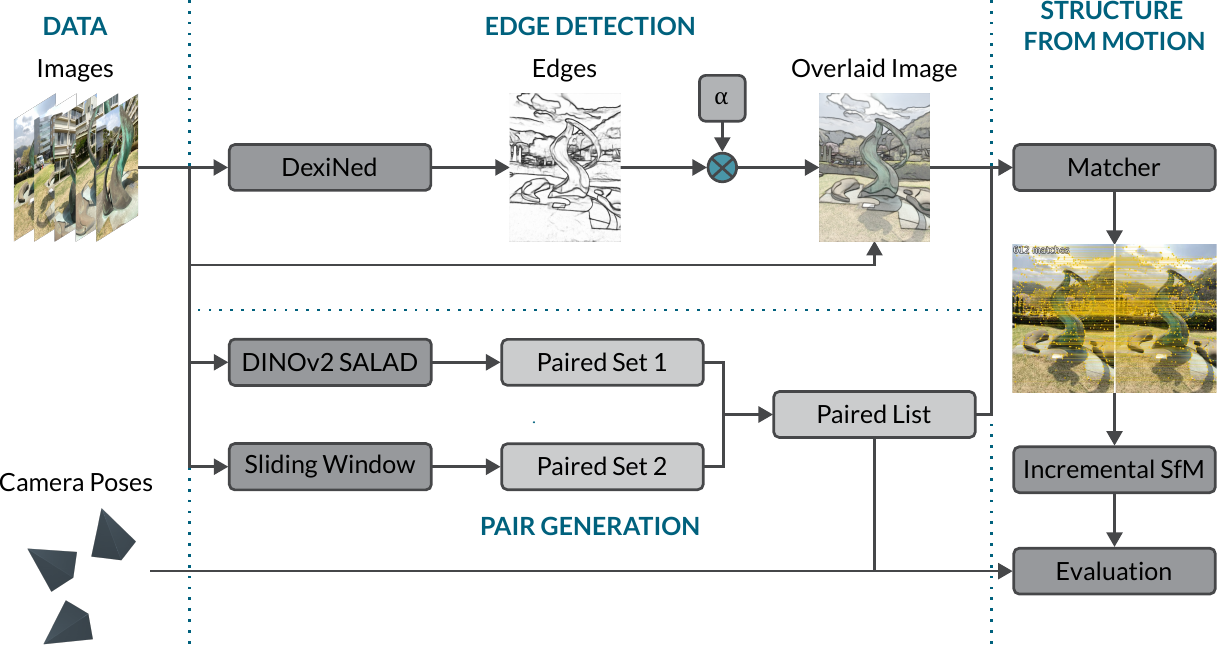}
    \caption{Flowchart illustrating our experiment pipeline with the edge detection, pair generation, and structure from motion steps.}
    \label{fig:flowchart}
\end{figure}

To perform our analysis, we employ a pipeline that is applied to each image matching method under comparison. The pipeline, as shown in~\Cref{fig:flowchart}, comprises 3 main sections: (1) Edge Extraction, (2) Pair Generation, and (3) \ac{sfm}. 

\paragraph{Edge Detection:} Images $I=\{I_n | 1\leqslant n\leqslant N\}$ are pre-processed by extracting edge maps using \ac{dexined}\cite{dexined_ext2023, dexined2020}, a learning-based edge detection method. Since \ac{dexined} outputs maps at different scales, we averaged them to create the edge images $I_E=\{I_{En} | 1\leqslant n\leqslant N\}$. This allows us to analyze the impact of highlighting edges with various image matching methods. For a deeper analysis, we include a variable $\alpha$, representing the blending factor between $I$ and $I_E$. Specifically, when $\alpha=0$, $I$ is used as input to the image matcher, when $\alpha=1$, $I_E$ is used and when $0 < \alpha < 1$, a blend $I_{\alpha}$ of $I$ and $I_E$ is used. 

\paragraph{Image Pair Generation:} To select the image pairs for matching, we first pair sequential frames (sliding window method). However, since not all sequences are sequential, we also employ the pre-trained model from DINOv2-SALAD\cite{Izquierdo_CVPR_2024_SALAD} to generate additional pairs. DINOv2-SALAD is a Visual Place Recognition method that finds image correspondences from a database to a query image. DINOv2 is used as the backbone for local feature extraction and SALAD is used as an aggregation technique that leverages the relationship between features and learned clusters using the optimal transport method to generate global features. 

\paragraph{Structure-from-motion:} Images $I$, $I_E$, or $I_{\alpha}$ from the paired list are fed to the image matcher. The resulting matched features and descriptors are fed into the incremental \ac{sfm} block, which we adopt from COLMAP\cite{schoenberger2016sfm,schoenberger2016mvs}. This block first ensures only geometrically consistent matches are retained. Then an incremental reconstruction process is initialized with an image pair, after which new images are sequentially registered and triangulated into \ac{3d} points. Within this process, bundle adjustment and \ac{ransac}\cite{fischler1981random} are used for pose estimation and outlier rejection. The output of this process includes the estimated poses, which are then evaluated using the ground truth.

\section{Experiments}
\label{sec:experiments}

The methods selected for comparison are, in alphabetical order, ALIKED with LightGlue (referred to as LG), ALIKED with LG and Keypt2Subpx (K2S), DeDoDe with \ac{dsm}, DeDoDe with LG, DeDoDe with LG and K2S, DeDoDe with Steerers, DISK with LG, Efficient-LoFTR (eLoFTR), LoFTR, MASt3R, RoMa, SIFT with LG, SIFT with \ac{nn}, SuperPoint (SP) with LG and GIM, SP with OmniGlue (OG), SP with SuperGlue (SG), xFeat with LG, xFeat with LG and K2S, xFeat with \ac{mnn}, and xFeat with \ac{mnn} and K2S.

DeDoDe, DISK, LoFTR, MASt3R, RoMa and SG were selected following the Niantic's Map-free Relocalization Challenge Evaluation Leaderboard.\footnote{\href{https://research.nianticlabs.com/mapfree-reloc-benchmark/leaderboard?t=single&f=all}{Map-free Relocalization Challenge Evaluation Leaderboard}} We selected eLoFTR, GIM, K2S, MASt3R, OG, RoMa, Steerers, and XFeat for their novelty, as they have been proposed in 2024. We excluded DUSt3R and included its more successful successor, MASt3R, as shown in experiments in \cite{leroy2024grounding}. Following the pipeline in~\Cref{fig:flowchart}, we conducted experiments to explore the limitations and advantages of the selected image matching methods.

\subsection{Datasets}

The selected image matching methods report having been trained on over 20 different datasets. However, these datasets predominantly feature outdoor scenes captured during the daytime. In addition, the images, whether synthetic or real, are typically collected using sequential frames with consistent camera intrinsics. In contrast, the \ac{imc24}\cite{image-matching-challenge-2024} dataset presents numerous challenges such as phototourism and historical preservation, night versus day, illumination changes, temporal and seasonal changes, rotations, multiple camera intrinsics, nonsequential scenes, aerial and mixed aerial-ground, repeated structures, natural environments, and transparencies and reflections.

Most datasets used to train our selected models typically include only one or two of these challenges, which classifies them as in-domain datasets. In comparison, datasets that encompass many of these challenges are considered out-of-domain. Among the many datasets used for training, the Aachen dataset~\cite{sattler2018benchmarking} is the closest to \ac{imc24}, as it includes seasonal changes, different camera intrinsics, and both daytime and nighttime scenes. Notably, only the ALIKED model was trained on this dataset. Given this distinction, our evaluation includes both in-domain and out-of-domain datasets (\Cref{fig:datasets}) with ground truth poses:

\begin{figure}[!t]
    \centering
    \includegraphics[width=\linewidth]{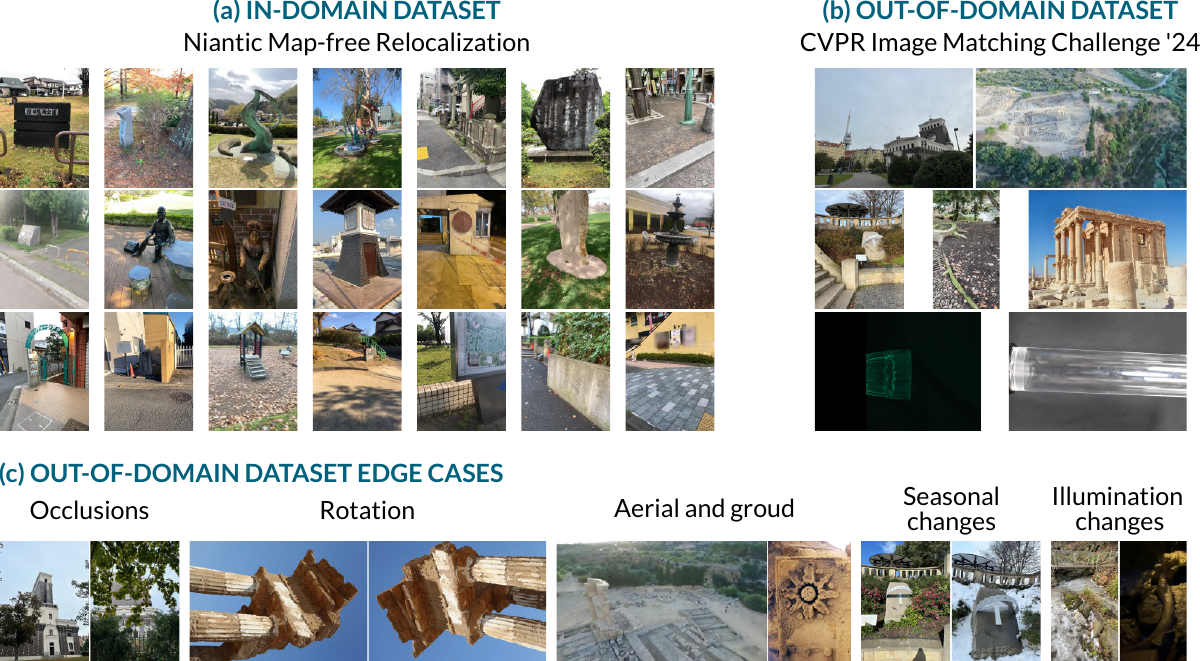}
    \caption{(a) In-domain Niantic dataset (21 scenes from validation dataset)\cite{arnold2022map}. (b) Out-of-domain \ac{imc24} dataset (7 scenes from training data)\cite{image-matching-challenge-2024}. (c) Edge cases from the out-of-domain dataset including occlusions, rotations, aerial and ground acquisitions, seasonal, and illumination changes.}
    \label{fig:datasets}
\end{figure}

\begin{enumerate}
    \item In-domain (\Cref{fig:datasets}(a)): The Niantic Map-free Relocalization Dataset\cite{arnold2022map} includes 655 outdoor places, each containing a small scene of interest like a sculpture or sign. All images are captured using a single camera during daytime in a sequential fashion with view overlap between images. Some frames exhibit significant movement and noise and not all frames contain the object of interest. We use their validation scenes with ground truth poses to evaluate the models. Due to computational constraints, we use the first 21 out of the 65 scenes. Although this dataset was originally designed for Map-Free relocalization using only one photo as reference per scene for camera pose regression, we utilize it to match multiple images from different viewpoints within the same scene.
    \item Out-of-domain (\Cref{fig:datasets}(b)): The Image Matching Challenge 2024 Dataset\cite{image-matching-challenge-2024} (IMC24) includes 7 scenes with multiple distinct challenges (\Cref{fig:datasets}(c)). This dataset was created for the task we are evaluating against. 
\end{enumerate}
To address computational constraints, we limit each scene to a maximum of 200 images. If a scene contains more than 200 images, we uniformly sub-sample them to approximately meet this limit. All images were resized to 1024x1024.

\subsection{Experimental Setup}

All experiments were conducted on a workstation equipped with an AMD Ryzen Threadripper PRO 5975WX CPU and NVIDIA RTX\textsuperscript{TM} A6000 GPUs. We perform two experiments following the pipeline detailed in \Cref{sec:methodology}. In the first experiment, we test the various methods without using \ac{dexined} (images $I$ are used, $\alpha=0$). In the second experiment, we use the pre-trained \ac{dexined} model to generate edge images $I_{E}$ and use them as input to our matchers ($\alpha=1$).

In the pipeline's DINOv2-SALAD pair generation, we selected the 10 closest images (based on Euclidean distance) to each image in the \ac{imc24} dataset and the 5 closest images in the Niantic dataset due to its sequential nature. All matcher parameters were kept at their default settings as described in\cite{Berton_2024_EarthMatch}.\footnote{\href{https://github.com/gmberton/EarthMatch}{https://github.com/gmberton/EarthMatch}} Improved results could have been achieved by tuning these parameters for each specific scene; to ensure fairness, adjustments were not made. For the incremental mapping block, we set the following requirements: a minimum of 3 successfully registered images per model, a minimum of 5 matches between image pairs, and a maximum of 2 separately registered models per scene.

\subsection{Metrics}
We evaluate the selected methods across three metrics: 
\begin{enumerate}
    \item Number of registered images $(N_{img})$: This metric represents the number of registered pairs after processing through our pipeline in \Cref{sec:methodology}.
    \item Average matching time $(Time)$: This is the average time taken to match a single pair for a specific method.
    \item Mean Average Accuracy $(mAA)$: We utilize the $mAA$ metric, which is widely recognized in the field and was first introduced by Jin \etal \cite{jin2021image}. Specifically, we adopt the variant used in the CVPR Image Matching Challenge 2023 (IMC23)\footnote{\href{https://www.kaggle.com/competitions/image-matching-challenge-2023}{IMC23 Competition Page}}. Although there have been modifications to the $mAA$ metric for camera pose estimation from the original paper and in the subsequent challenges (\ac{imc24}), our evaluation adheres to the 2023 version of the metric. This decision is based on consistency since we use their data and the same task. Also the \ac{imc24} metric introduces an extra RANSAC-like estimation step that adds a level of computational expense and randomness. To compute $mAA$, we first calculate the relative pose errors (rotation and translation) for the registered image pairs. These errors are then thresholded to determine the percentage of accurately registered pairs $A$; pairs are considered accurately registered if both rotation and translation errors are below the threshold. Images that do not get successfully registered are set to identity, effectively contributing a maximal error relative to any non-identity ground truth, which heavily penalizes the $mAA$ if not within acceptable thresholds. This process is performed across multiple thresholds. $A$ is then averaged over all thresholds to obtain the Average Accuracy $AA$. Finally, $mAA$ is calculated by averaging $AA$ over all scenes. 
\end{enumerate}

For the $mAA$ computations, we use five thresholds: $[0.25,0.5,1,2,5,10]$ degrees for rotation and $[0.025,0.05,0.1,0.2,0.5,1]$ meters for translation across all scenes, except for transparent object scenes. For these close-up scenes, we use smaller thresholds as recommended by \ac{imc24}: $[0.025,0.05,0.1,0.2,0.5,1]$ degrees for rotation and $[0.0025,0.005,0.01,0.02,0.05,0.1]$ meters for translation.

\section{Results \& Discussion}
 
\begin{figure}[!t]
    \centering
    \includegraphics[width=\linewidth]{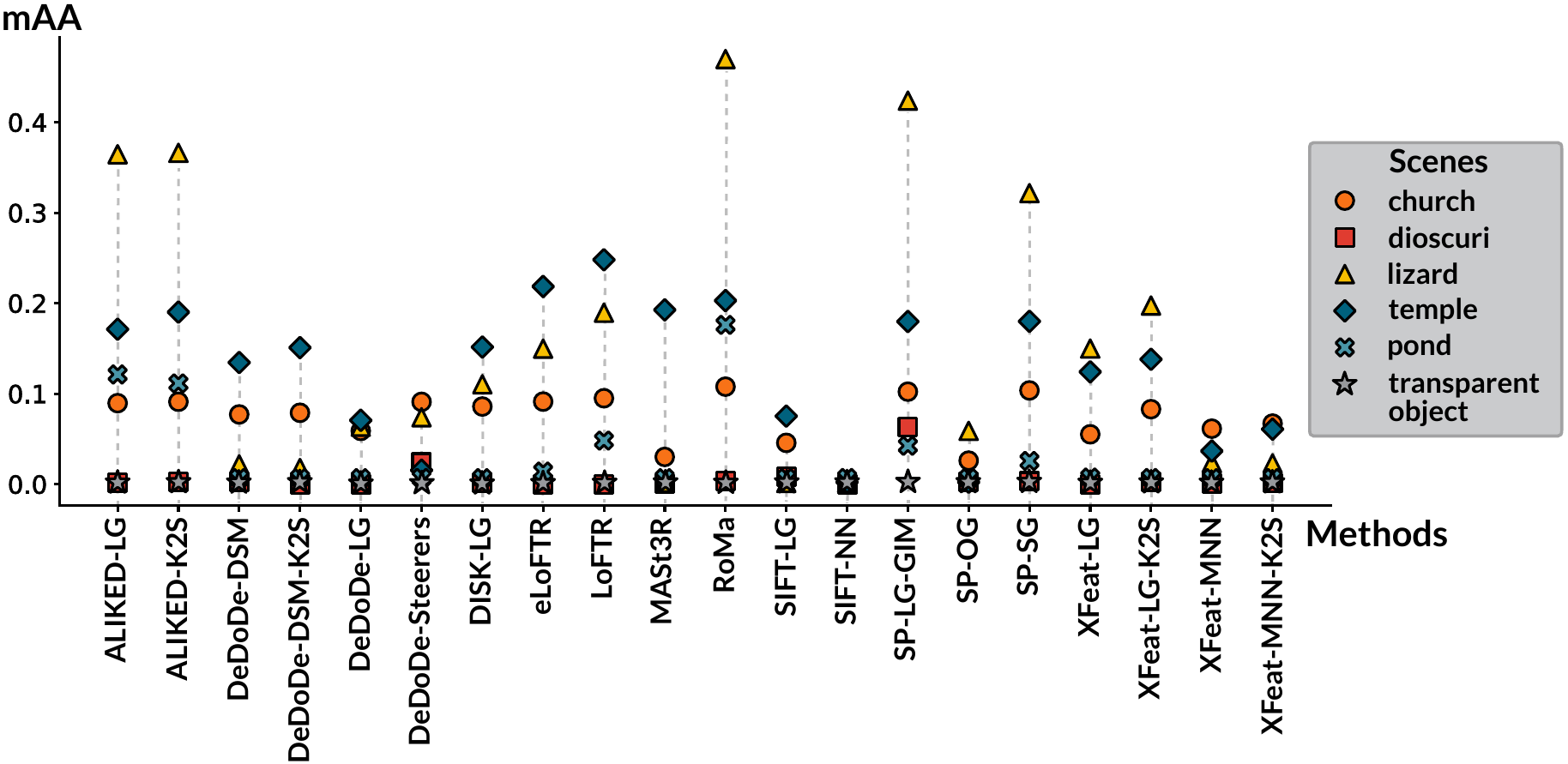}
    \caption{Visualizing $mAA$ for image matching methods on the \ac{imc24} scenes that represent various challenging categories.}
    \label{fig:challenge-plot}
\end{figure}

We present our main findings in \Cref{tab:results}, summarizing the performance metrics across two datasets without the use of edge detection. For the \ac{imc24} dataset, the $mAA$ was calculated for all scenes, with the exception of the transparent object scenes where no methods achieved a non-zero $mAA$. We also provide the $mAA$ results separated by the challenging scenes for \ac{imc24} in \Cref{fig:challenge-plot}.
\begin{table}[!t]
\centering
\caption{Comparison of image matching methods based on the mean average accuracy, $mAA$, the number of registered pairs, $N_{img}$, and the average time taken to match a single pair of images, $Time$, across two datasets: \ac{imc24} and Niantic. Best in \textbf{bold}, second and third-best \underline{underlined}, and handcrafted methods in \textit{italic}.}
\label{tab:results}
\resizebox{\linewidth}{!}{%
\begin{tabular}{cccccccccc} 
\toprule
\multirow{2}{*}{} & \multirow{2}{*}{\begin{tabular}[c]{@{}c@{}}Feature\\Extractor\end{tabular}} & \multirow{2}{*}{Matcher} & \multirow{2}{*}{Framework} & \multicolumn{3}{c}{IMC24} & \multicolumn{3}{c}{Niantic} \\ 
\cmidrule(r){5-10}
&  &  &  & $mAA\uparrow$ & $N_{img} (\%)\uparrow$ & $Time (s)\downarrow$ & $mAA\uparrow$ & $N_{img} (\%)\uparrow$ & $Time (s)\downarrow$\\
\midrule
\noalign{\smallskip}
\multirow{16}{*}{\rotatebox[origin=c]{90}{Detector-based}}
& ALIKED\cite{zhao2023aliked} & LG\cite{Lindenberger_2023_ICCV} &  & 0.148 & \uline{64.060} & 0.054 & 0.495 & \underline{88.533} & 0.055 \\
& ALIKED\cite{zhao2023aliked} & LG\cite{Lindenberger_2023_ICCV} & K2S\cite{kim2024learningmakekeypointssubpixel} & \underline{0.151} & 63.923 & 0.101 & 0.484 & 81.675 & 0.186 \\
& DeDoDe\cite{edstedt2023dedodedetectdont} & \textit{DSM} &  & 0.047 & 28.532 & 0.428  & 0.479 & 83.443 & 0.426 \\
& DeDoDe\cite{edstedt2023dedodedetectdont} & \textit{DSM} & K2S\cite{kim2024learningmakekeypointssubpixel} & 0.050  & 27.435 & 0.446 & 0.430 & 80.221 & 0.508 \\
& DeDoDe\cite{edstedt2023dedodedetectdont} & LG\cite{Lindenberger_2023_ICCV} &  & 0.039 & 32.510 & 0.316 & 0.454 & 80.849 & 0.318 \\
& DeDoDe\cite{edstedt2023dedodedetectdont} & Steerers\cite{bokman2024steerers} &  & 0.041 & 35.665 & 0.277 & 0.421 & 69.068 & 0.277 \\ 
& DISK\cite{tyszkiewicz2020disk} & LG\cite{Lindenberger_2023_ICCV} &  & 0.070 & 41.015 & 0.262 & 0.480 & 86.054 & 0.262 \\
& \textit{SIFT}\cite{lowe1999object} & LG\cite{Lindenberger_2023_ICCV} &  & 0.026 & 22.497 & 0.296 & 0.439 & 77.726  & 0.312 \\
& \textit{SIFT}\cite{lowe1999object} & \textit{NN} &  & 0.001 & 4.664 & 0.300 & 0.363 & 62.987 &  0.316 \\
& SP\cite{detone2018superpoint} & LG\cite{Lindenberger_2023_ICCV} & GIM\cite{shen2024gim} &  \underline{0.161} & \uline{68.861} & 0.070 & \textbf{0.560} & \textbf{88.698} & 0.067 \\
& SP\cite{detone2018superpoint} & OG\cite{jiang2024omniglue} &  & 0.018 & 25.377 & 1.641 & 0.457 & 75.644 & 1.881  \\
& SP\cite{detone2018superpoint} & SG\cite{sarlin2020superglue} &  & 0.126 & 56.927 & 0.067 &  \underline{0.532} & \underline{86.104} &\underline{0.054}\\
& XFeat\cite{potje2024xfeat} & LG\cite{Lindenberger_2023_ICCV} &  & 0.066  & 46.228 & \underline{0.023} &  \underline{0.501} & 85.592 & \underline{0.023} \\
 & XFeat\cite{potje2024xfeat} & LG\cite{Lindenberger_2023_ICCV} & K2S\cite{kim2024learningmakekeypointssubpixel} & 0.084 & 45.679 & \underline{0.044} & 0.476 & 82.105 & 0.095 \\
 & XFeat\cite{potje2024xfeat} & \textit{MNN} &  & 0.025 &25.652 & \textbf{0.014} & 0.388 & 68.440 & \textbf{0.014} \\
 & XFeat\cite{potje2024xfeat} & \textit{MNN} & K2S\cite{kim2024learningmakekeypointssubpixel} & 0.031& 25.926 & 0.072 & 0.448 & 78.470 & 0.107 \\
\noalign{\smallskip}
\midrule 
\noalign{\smallskip}
\multirow{4}{*}{\rotatebox[origin=c]{90}{Detector-free\hspace{-0.16cm}}}
& \multicolumn{2}{c}{eLoFTR\cite{wang2024efficient}} &  & 0.094 & 48.834 & 0.135  & 0.393 & 75.826 & 0.147 \\
& \multicolumn{2}{c}{LoFTR\cite{sun2021loftr}} &  & 0.115  & 54.733 & 0.210 & 0.480 & 84.534 & 0.285 \\
 & \multicolumn{2}{c}{MASt3R\cite{leroy2024grounding}} &  & 0.045 & 23.045 & 3.433 & 0.243 & 70.258 & 3.553 \\
 & \multicolumn{2}{c}{RoMa\cite{edstedt2024roma}} &  & \textbf{0.191} & \textbf{73.114} & 0.482 & 0.104 & 15.367 & 0.479 \\
\noalign{\smallskip}
\bottomrule
\end{tabular}
}
\end{table}

\paragraph{Average matching time (Time):}In our results, the fastest method is XFeat-MNN followed by XFeat-LG and XFeat-LG-K2S for \ac{imc24} and similarly, XFeat-MNN, XFeat-LG, and SP-SG for Niantic. This suggests that XFeat is a time-efficient method that aligns with their lightweight architecture contribution. Whereas, the slowest method is MASt3R followed by SP-OG. It is also worth mentioning that even though dense matchers, like LoFTR and eLoFTR, have comparable average matching time to detector-based methods, they take significantly longer in the incremental \ac{sfm} block. This is a direct consequence of the high number of matches generated from these matchers due to their dense nature. ALIKED-LG and SP-LG-GIM perform very well on time for both datasets.

\paragraph{Out-of-Domain Challenges:} Within the out-of-domain sequences, some are particularly challenging. From our results in \Cref{fig:challenge-plot}, we see that all methods fail to register images with the transparent sequences at the default hyperparameters. This leads us to believe that this problem is still entirely unsolved by current solutions. Other difficult sequences include Dioscuri, containing 90$^{\circ}$ and 180$^{\circ}$ flips, and pond, containing night and day scenes with occlusions. For Dioscuri, SP-LG-GIM and DeDoDe-Steerers methods get slightly better results than the other methods. This aligns with Steerers' rotation invariance claim and GIM’s extensive rotation augmentations during training. As for the pond sequence, only ALIKED and RoMa methods were able to handle them. This could be because ALIKED was trained on a dataset containing night and day scenes. In general, results are quite low for these challenging scenes, as also seen in the \ac{imc24} Leaderboard which suggests ample opportunity for future contributions.
We notice that the percentage gap between the out-of-domain and in-domain experiments varies significantly among methods. For both $mAA$ and $N_{img}$, the lowest gaps are with ALIKED-LG-K2S, SP-LG-GIM, and ALIKED-LG, which seems reasonable given that ALIKED is trained on more challenging data than other methods. It also suggests that GIM can boost SP-LG's generalizability. Some of the biggest gaps showing lower generalizability come from OmniGlue and SIFT methods. SIFT could be explained by its geometrical features that are harder to find in challenging images.

\paragraph{Surprising Results:} The performance of MASt3R on the Niantic dataset is comparatively poor when evaluated against the map-free Niantic leaderboard. One reason for the performance discrepancy is that it is a different task than image matching across a scene. MASt3R was designed and trained for a different task, map-free relocalization, whereas the task in this paper involves matching many frames across varying camera intrinsics. This task requires that keypoints remain consistent not just across two images but across a whole dataset, otherwise they would be removed on geometric verification. In addition, MASt3R is trained on very different image resolution and aspect ratio than those that are used for inference in this study. In terms of matching times, MASt3R ranks last, even though \cite{leroy2024grounding} notes that their fast reciprocal matching scheme accelerates matching. We suggest that it may be because we have implemented MASt3R on dataset with higher resolution, which is mentioned in \cite{leroy2024grounding} to require significantly higher computing time. Another reason is perhaps that their setup is better. 

Another unexpected finding was that RoMa does quite well on the \ac{imc24} dataset, but not on Niantic. In comparison to the Niantic leaderboard, RoMa is always paired with Mickey \cite{barrosolaguna2024matching2dimages3d}, which is reported to achieve state-of-the-art performance on Map-Free Relocalization, or DPT-KITTI \cite{ranftl2021vision}, although the major performance gap between the two datasets warrants further investigation. 

\begin{figure}[!t]
    \centering
    \includegraphics[width=\linewidth]{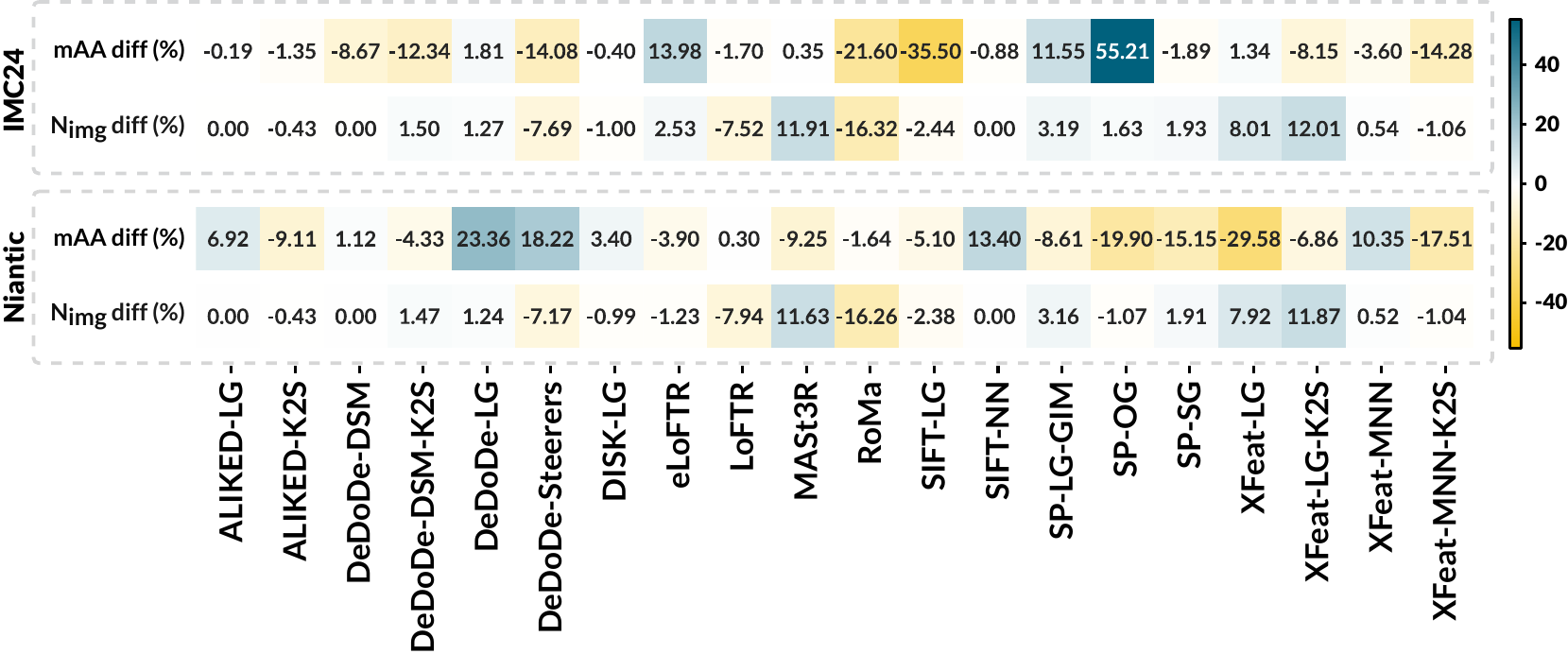}
    \caption{Heatmap representing the image matching models' performance changes when incorporating edge detection.}
    \label{fig:edge-detection}
\end{figure}

\paragraph{Edge Detection:} Edge detection as a pre-processing step shows mixed effects on our evaluation metrics; this is visually represented in \Cref{fig:edge-detection}. The heatmap reports the percentage difference in $mAA$ and $N_{img}$ with and without edge detection. 
When using edge detection $(\alpha=1)$ on the Niantic dataset, $mAA$ improves with SIFT-NN and makes DeDoDe-LG the best-performing model. Although also using LG as a matcher, SIFT-LG performs worse across the board using edge detection. Similarly, edge detection does not benefit RoMa. Specifically, on the \ac{imc24} dataset, the most generalizable method across our experiments, SP-LG-GIM was improved with edge detection. Despite the large percent improvement shown for SP-OG, it is not significant due to the low original $mAA$ value, which even when improved (55.21\%) would still be very small and likely attributable to randomness. Note that since statistical variability increases with smaller sample sizes of a population, \ac{imc24} observations are less trustworthy than the Niantic dataset. This problem is further intensified by an inherently more variable underlying distribution in \ac{imc24}. 

When using an $\alpha=0.5$, we notice that $mAA_{\alpha=0.5}$ values are either very similar to $mAA_{\alpha=1}$ or lie in between $mAA_{\alpha=0}$ and $mAA_{\alpha=1}$. Interestingly, with $\alpha=0.5$, some methods improved, including XFeat-LG with \ac{imc24}, and ALIKED-LG-K2S, DeDoDe-LG-K2S, and RoMa with Niantic. Thus we recommend investigating different alpha values when using these specific methods.

In general, edge detection has a varied effect. We suspect it is related to how much the method relies on image edges versus other types of features. This could be affected by the method's architecture, training data, and augmentation strategies. Finally, it might be worth investigating the effect of different pre-processing options such as cropping to areas of overlap, correcting for rotations, or augmenting training with steps like this. 

\paragraph{Metric Ambiguity:} 
In the context of evaluating camera pose accuracy, the $mAA$ and AUC metrics, commonly reported, warrant a closer examination, especially in scenarios where a low number of camera poses are registered. 
The commonly reported image matching metric $mAA$ often lacks clarity regarding the specific errors computed, the thresholds applied, and the treatment of unregistered images. The benchmarking paper \cite{jin2021image} introduced this widely used baseline in 2021 which recommends setting the pose errors for unregistered images to infinity; many papers report using this metric. This approach can significantly distort statistical measures like the mean and thus can preclude meaningful comparisons. In contrast, the 2023 Image Matching Challenge iteration modified $mAA$ to use identity matrices for unregistered images, which avoids infinite distortions and maintains all errors within a finite range. However, this approach can underrepresent the severity of mismatches when the ground truth is close to the identity transformation. This method also changes the relative translation error from angular, used in \cite{jin2021image}, to Euclidean-distance-based. 

When the number of registered poses is low, the $mAA$ metric effectively becomes a gauge of registration success rate, rather than solely a measure of pose accuracy. $mAA$ might suggest low accuracy not because of poor pose estimations but because a minimal number of images were successfully registered. This conflation of registration success with pose accuracy creates a scenario where a low $mAA$ value could indicate two fundamentally different issues. The description of the metric is \textit{mismatched} with its actual informational content, misleading interpretations of method performance. Knowing the number of successfully registered poses is important to interpret these results accurately.

We recommend (1) researchers consistently report which version of the metric they are using, since without clarity on how unregistered images are handled, comparing $mAA$ values between papers can be misleading. (2) Given the inherent limitations of this metric, it is beneficial to also separately report the number of images successfully registered to provide a clearer evaluation.

\section{Conclusion}
\label{sec:conclusion}
In this paper, we used SfM to evaluate various image matching techniques. Our key findings are: (1) Few methods are generalizable enough to handle out-of-domain challenges, with none successfully registering transparent object scenes, leaving this an open problem. (2) The current $mAA$ metric is ambiguous in error definition and treatment of unregistered images, and under certain conditions, there is a misalignment between the metric's intended meaning and content. Finally, (3) the poor evaluation results across all methods on the \ac{imc24} dataset show the need for datasets that capture a wider range of variability. Larger datasets that better approximate real-world distributions are still needed for developing methods that generalize well. Given these challenges, choosing the best overall method is context-dependent and we hope this evaluation serves as a guide for researchers selecting and developing image matching methods.

\section*{Acknowledgements}
This work was supported in whole, or in part, by the Wellcome/EPSRC Centre for Interventional and Surgical Sciences (WEISS) [203145/Z/16/Z], the Department of Science, Innovation and Technology (DSIT), the Royal Academy of Engineering under the Chair in Emerging Technologies programme, Engineering and Physical Sciences Research Council (EPSRC) [EP/W00805X/1, EP/Y01958X/1, EP/P012841/1]; Horizon 2020 FET [GA863146]. Sierra Bonilla is supported by the UKRI AI Centre for Doctoral Training in Foundational Artificial Intelligence (FAI CDT) [EP/S021566/1]. Rema Daher is supported by the UCL Centre for Digital Innovation through the Amazon Web Services (AWS) Doctoral Scholarship in Digital Innovation 2023/2024. 
For the purpose of open access, the authors have applied a CC BY public copyright licence to any author accepted manuscript version arising from this submission.

%
%
\bibliographystyle{splncs04}
\bibliography{main}

\begin{thebibliography}{10}
\providecommand{\url}[1]{\texttt{#1}}
\providecommand{\urlprefix}{URL }
\providecommand{\doi}[1]{https://doi.org/#1}

\bibitem{antoun2020towards}
Antoun, M., Asmar, D., Daher, R.: Towards richer 3d reference maps in urban scenes. In: 2020 17th Conference on Computer and Robot Vision (CRV). pp. 39--45. IEEE Computer Society (2020). \doi{10.1109/CRV50864.2020.00014}

\bibitem{arnold2022map}
Arnold, E., Wynn, J., Vicente, S., Garcia-Hernando, G., Monszpart, A., Prisacariu, V., Turmukhambetov, D., Brachmann, E.: Map-free visual relocalization: Metric pose relative to a single image. In: European Conference on Computer Vision. pp. 690--708. Springer (2022)

\bibitem{bagnolo2019hbim}
Bagnolo, V., Argiolas, R., Cuccu, A.: Hbim for archaeological sites: from sfm based survey to algorithmic modeling. The International Archives of the Photogrammetry, Remote Sensing and Spatial Information Sciences  \textbf{42},  57--63 (2019). \doi{10.5194/isprs-archives-XLII-2-W9-57-2019}

\bibitem{barrosolaguna2024matching2dimages3d}
Barroso-Laguna, A., Munukutla, S., Prisacariu, V.A., Brachmann, E.: Matching 2d images in 3d: Metric relative pose from metric correspondences. In: Proceedings of the IEEE/CVF Conference on Computer Vision and Pattern Recognition. pp. 4852--4863 (2024). \doi{10.48550/arXiv.2404.06337}

\bibitem{bayherbert}
Bay, H., Tuytelaars, T., Van~Gool, L.: Surf: Speeded up robust features. In: Computer Vision-ECCV 2006. vol.~3951, pp. 404--417 (07 2006). \doi{10.1007/11744023_32}

\bibitem{image-matching-challenge-2024}
Bellavia, F., Mishkin, D., Matas, J., Morelli, L., Remondino, F., Sun, W., Tabb, A., Trulls, E., Yi, K.M., Dane, S., Chow, A.: Image matching challenge 2024 - hexathlon (2024), \url{https://kaggle.com/competitions/image-matching-challenge-2024}

\bibitem{Berton_2024_EarthMatch}
Berton, G., Goletto, G., Trivigno, G., Stoken, A., Caputo, B., Masone, C.: Earthmatch: Iterative coregistration for fine-grained localization of astronaut photography. In: Proceedings of the IEEE/CVF Conference on Computer Vision and Pattern Recognition. pp. 4264--4274 (2024)

\bibitem{bokman2024steerers}
B{\"o}kman, G., Edstedt, J., Felsberg, M., Kahl, F.: Steerers: A framework for rotation equivariant keypoint descriptors. In: Proceedings of the IEEE/CVF Conference on Computer Vision and Pattern Recognition. pp. 4885--4895 (2024)

\bibitem{borhani2019digital}
Borhani, N., Bower, A.J., Boppart, S.A., Psaltis, D.: Digital staining through the application of deep neural networks to multi-modal multi-photon microscopy. Biomedical optics express  \textbf{10}(3),  1339--1350 (2019). \doi{10.1364/BOE.10.001339}

\bibitem{calonder2010brief}
Calonder, M., Lepetit, V., Strecha, C., Fua, P.: Brief: Binary robust independent elementary features. In: Computer Vision--ECCV 2010: 11th European Conference on Computer Vision, Heraklion, Crete, Greece, September 5-11, 2010, Proceedings, Part IV 11. pp. 778--792. Springer (2010). \doi{10.1166/ASL.2017.7336}

\bibitem{detone2018superpoint}
DeTone, D., Malisiewicz, T., Rabinovich, A.: Superpoint: Self-supervised interest point detection and description. In: Proceedings of the IEEE conference on computer vision and pattern recognition workshops. pp. 224--236 (2018). \doi{10.1109/CVPRW.2018.00060}

\bibitem{edstedt2023dedodedetectdont}
Edstedt, J., B{\"o}kman, G., Wadenb{\"a}ck, M., Felsberg, M.: Dedode: Detect, don’t describe — describe, don’t detect for local feature matching. 2024 International Conference on 3D Vision (3DV) pp. 148--157 (2023). \doi{10.1109/3DV62453.2024.00035}

\bibitem{edstedt2024roma}
Edstedt, J., Sun, Q., B{\"o}kman, G., Wadenb{\"a}ck, M., Felsberg, M.: Roma: Robust dense feature matching. In: Proceedings of the IEEE/CVF Conference on Computer Vision and Pattern Recognition. pp. 19790--19800 (2024)

\bibitem{eppstein1998raising}
Eppstein, D., Erickson, J.: Raising roofs, crashing cycles, and playing pool: Applications of a data structure for finding pairwise interactions. Discrete \& Computational Geometry  \textbf{22},  569--592 (1998). \doi{10.1145/276884.276891}

\bibitem{fischler1981random}
Fischler, M.A., Bolles, R.C.: Random sample consensus: a paradigm for model fitting with applications to image analysis and automated cartography. Communications of the ACM  \textbf{24}(6),  381--395 (1981). \doi{10.1145/358669.358692}

\bibitem{furukawa2009accurate}
Furukawa, Y., Ponce, J.: Accurate, dense, and robust multiview stereopsis. IEEE Transactions on Pattern Analysis and Machine Intelligence  \textbf{32},  1362--1376 (2010). \doi{10.1109/TPAMI.2009.161}

\bibitem{geiger2012we}
Geiger, A., Lenz, P., Urtasun, R.: Are we ready for autonomous driving? the kitti vision benchmark suite. In: 2012 IEEE conference on computer vision and pattern recognition. pp. 3354--3361. IEEE (2012). \doi{10.1109/CVPR.2012.6248074}

\bibitem{geiger2011stereoscan}
Geiger, A., Ziegler, J., Stiller, C.: Stereoscan: Dense 3d reconstruction in real-time. In: 2011 IEEE Intelligent Vehicles Symposium (IV). pp. 963--968 (2011). \doi{10.1109/IVS.2011.5940405}

\bibitem{hane2011stereo}
H{\"a}ne, C., Zach, C., Lim, J., Ranganathan, A., Pollefeys, M.: Stereo depth map fusion for robot navigation. In: 2011 IEEE/RSJ International Conference on Intelligent Robots and Systems. pp. 1618--1625. IEEE (2011). \doi{10.1109/IROS.2011.6094704}

\bibitem{iheaturu2022simplified}
Iheaturu, C., Okolie, C., Ayodele, E., Egogo-Stanley, A., Musa, S., Speranza, C.I.: A simplified structure-from-motion photogrammetry approach for urban development analysis. Remote sensing applications: Society and Environment  \textbf{28} (2022). \doi{10.1016/j.rsase.2022.100850}

\bibitem{Izquierdo_CVPR_2024_SALAD}
Izquierdo, S., Civera, J.: Optimal transport aggregation for visual place recognition. In: Proceedings of the IEEE/CVF Conference on Computer Vision and Pattern Recognition. pp. 17658--17668 (2024)

\bibitem{jiang2024omniglue}
Jiang, H., Karpur, A., Cao, B., Huang, Q., Araujo, A.: Omniglue: Generalizable feature matching with foundation model guidance. In: Proceedings of the IEEE/CVF Conference on Computer Vision and Pattern Recognition. pp. 19865--19875 (2024). \doi{10.48550/arXiv.2405.12979}

\bibitem{jin2021image}
Jin, Y., Mishkin, D., Mishchuk, A., Matas, J., Fua, P., Yi, K.M., Trulls, E.: Image matching across wide baselines: From paper to practice. International Journal of Computer Vision  \textbf{129}(2),  517--547 (2021). \doi{10.1007/s11263-020-01385-0}

\bibitem{juan1982programme}
Juan, J.: Programme de classification hi{\'e}rarchique par l'algorithme de la recherche en cha{\^\i}ne des voisins r{\'e}ciproques. Les cahiers de l'analyse des donn{\'e}es  \textbf{7}(2),  219--225 (1982)

\bibitem{khan2021vision}
Khan, A., Mineo, C., Dobie, G., Macleod, C., Pierce, G.: Vision guided robotic inspection for parts in manufacturing and remanufacturing industry. Journal of Remanufacturing  \textbf{11}(1),  49--70 (2021). \doi{10.1007/s13243-020-00091-x}

\bibitem{kim2024learningmakekeypointssubpixel}
Kim, S., Pollefeys, M., Barath, D.: Learning to make keypoints sub-pixel accurate (2024), \url{https://arxiv.org/abs/2407.11668}

\bibitem{leroy2024grounding}
Leroy, V., Cabon, Y., Revaud, J.: Grounding image matching in 3d with mast3r. arXiv preprint arXiv:2406.09756  (2024)

\bibitem{Lindenberger_2023_ICCV}
Lindenberger, P., Sarlin, P.E., Pollefeys, M.: Lightglue: Local feature matching at light speed. 2023 IEEE/CVF International Conference on Computer Vision (ICCV) pp. 17581--17592 (2023). \doi{10.1109/ICCV51070.2023.01616}

\bibitem{lowe1999object}
Lowe, D.G.: Object recognition from local scale-invariant features. In: Proceedings of the seventh IEEE international conference on computer vision. vol.~2, pp. 1150--1157. IEEE (1999). \doi{10.1109/ICCV.1999.790410}

\bibitem{oquab2023dinov2}
Oquab, M., Darcet, T., Moutakanni, T., Vo, H.V., Szafraniec, M., Khalidov, V., Fernandez, P., HAZIZA, D., Massa, F., El-Nouby, A., Assran, M., Ballas, N., Galuba, W., Howes, R., Huang, P.Y., Li, S.W., Misra, I., Rabbat, M., Sharma, V., Synnaeve, G., Xu, H., Jegou, H., Mairal, J., Labatut, P., Joulin, A., Bojanowski, P.: {DINO}v2: Learning robust visual features without supervision. Transactions on Machine Learning Research  (2024). \doi{10.48550/arXiv.2304.07193}

\bibitem{potje2024xfeat}
Potje, G., Cadar, F., Araujo, A., Martins, R., Nascimento, E.R.: Xfeat: Accelerated features for lightweight image matching. In: Proceedings of the IEEE/CVF Conference on Computer Vision and Pattern Recognition. pp. 2682--2691 (2024)

\bibitem{ranftl2021vision}
Ranftl, R., Bochkovskiy, A., Koltun, V.: Vision transformers for dense prediction. In: Proceedings of the IEEE/CVF international conference on computer vision. pp. 12179--12188 (2021). \doi{10.1109/ICCV48922.2021.01196}

\bibitem{rosten2006machine}
Rosten, E., Drummond, T.: Machine learning for high-speed corner detection. In: Computer Vision--ECCV 2006: 9th European Conference on Computer Vision, Graz, Austria, May 7-13, 2006. Proceedings, Part I 9. pp. 430--443. Springer (2006). \doi{10.1007/11744023_34}

\bibitem{rublee2011orb}
Rublee, E., Rabaud, V., Konolige, K., Bradski, G.: Orb: An efficient alternative to sift or surf. In: 2011 International conference on computer vision. pp. 2564--2571. IEEE (2011). \doi{10.1109/ICCV.2011.6126544}

\bibitem{sarlin2020superglue}
Sarlin, P.E., DeTone, D., Malisiewicz, T., Rabinovich, A.: Superglue: Learning feature matching with graph neural networks. In: Proceedings of the IEEE/CVF conference on computer vision and pattern recognition. pp. 4938--4947 (2020). \doi{10.1109/cvpr42600.2020.00499}

\bibitem{sattler2018benchmarking}
Sattler, T., Maddern, W., Toft, C., Torii, A., Hammarstrand, L., Stenborg, E., Safari, D., Okutomi, M., Pollefeys, M., Sivic, J., et~al.: Benchmarking 6dof outdoor visual localization in changing conditions. In: Proceedings of the IEEE conference on computer vision and pattern recognition. pp. 8601--8610 (2018). \doi{10.1109/CVPR.2018.00897}

\bibitem{schoenberger2016sfm}
Sch\"{o}nberger, J.L., Frahm, J.M.: Structure-from-motion revisited. In: Conference on Computer Vision and Pattern Recognition (CVPR) (2016). \doi{10.1109/CVPR.2016.445}

\bibitem{schoenberger2016mvs}
Sch\"{o}nberger, J.L., Zheng, E., Pollefeys, M., Frahm, J.M.: Pixelwise view selection for unstructured multi-view stereo. In: European Conference on Computer Vision (ECCV) (2016). \doi{10.1007/978-3-319-46487-9_31}

\bibitem{shen2024gim}
Shen, X., Cai, Z., Yin, W., M{\"u}ller, M., Li, Z., Wang, K., Chen, X., Wang, C.: Gim: Learning generalizable image matcher from internet videos. arXiv preprint arXiv:2402.11095  (2024)

\bibitem{dexined2020}
Soria, X., Riba, E., Sappa, A.: Dense extreme inception network: Towards a robust cnn model for edge detection. In: 2020 IEEE Winter Conference on Applications of Computer Vision (WACV). pp. 1912--1921 (2020). \doi{10.1109/WACV45572.2020.9093290}

\bibitem{dexined_ext2023}
Soria, X., Sappa, A., Humanante, P., Akbarinia, A.: Dense extreme inception network for edge detection. Pattern Recognition  \textbf{139},  109461 (2023). \doi{10.1016/j.patcog.2023.109461}

\bibitem{sun2021loftr}
Sun, J., Shen, Z., Wang, Y., Bao, H., Zhou, X.: Loftr: Detector-free local feature matching with transformers. In: Proceedings of the IEEE/CVF conference on computer vision and pattern recognition. pp. 8922--8931 (2021). \doi{10.1109/CVPR46437.2021.00881}

\bibitem{tyszkiewicz2020disk}
Tyszkiewicz, M., Fua, P., Trulls, E.: Disk: Learning local features with policy gradient. Advances in Neural Information Processing Systems  \textbf{33},  14254--14265 (2020)

\bibitem{ullman1979interpretation}
Ullman, S.: The interpretation of structure from motion. Proceedings of the Royal Society of London. Series B. Biological Sciences  \textbf{203}(1153),  405--426 (1979). \doi{10.1109/CVPR.2016.445}

\bibitem{voigtlaender2019mots}
Voigtlaender, P., Krause, M., Osep, A., Luiten, J., Sekar, B.B.G., Geiger, A., Leibe, B.: Mots: Multi-object tracking and segmentation. In: Proceedings of the ieee/cvf conference on computer vision and pattern recognition. pp. 7942--7951 (2019). \doi{10.1109/CVPR.2019.00813}

\bibitem{wang2024dust3r}
Wang, S., Leroy, V., Cabon, Y., Chidlovskii, B., Revaud, J.: Dust3r: Geometric 3d vision made easy. In: Proceedings of the IEEE/CVF Conference on Computer Vision and Pattern Recognition. pp. 20697--20709 (2024). \doi{10.48550/arXiv.2312.14132}

\bibitem{wang2024efficient}
Wang, Y., He, X., Peng, S., Tan, D., Zhou, X.: Efficient loftr: Semi-dense local feature matching with sparse-like speed. In: Proceedings of the IEEE/CVF Conference on Computer Vision and Pattern Recognition. pp. 21666--21675 (2024). \doi{10.48550/arXiv.2403.04765}

\bibitem{weinzaepfel2023crocoselfsupervisedpretraining3d}
Weinzaepfel, P., Leroy, V., Lucas, T., Br{\'e}gier, R., Cabon, Y., Arora, V., Antsfeld, L., Chidlovskii, B., Csurka, G., Revaud, J.: Croco: Self-supervised pre-training for 3d vision tasks by cross-view completion. Advances in Neural Information Processing Systems  \textbf{35},  3502--3516 (2022). \doi{48550/arXiv.2210.10716}

\bibitem{xiang2017posecnn}
Xiang, Y., Schmidt, T., Narayanan, V., Fox, D.: Posecnn: A convolutional neural network for 6d object pose estimation in cluttered scenes. ArXiv  \textbf{abs/1711.00199} (2017). \doi{10.15607/RSS.2018.XIV.019}

\bibitem{Xu_2024}
Xu, S., Chen, S., Xu, R., Wang, C., Lu, P., Guo, L.: Local feature matching using deep learning: A survey. Information Fusion  \textbf{107} (2024). \doi{10.1016/j.inffus.2024.102344}

\bibitem{yang20243d}
Yang, Z., Dai, J., Pan, J.: 3d reconstruction from endoscopy images: A survey. Computers in Biology and Medicine p. 108546 (2024). \doi{10.1016/j.compbiomed.2024.108546}

\bibitem{yi2016lift}
Yi, K.M., Trulls, E., Lepetit, V., Fua, P.: Lift: Learned invariant feature transform. In: Computer Vision--ECCV 2016: 14th European Conference, Amsterdam, The Netherlands, October 11-14, 2016, Proceedings, Part VI 14. pp. 467--483. Springer (2016). \doi{10.1007/978-3-319-46466-4_28}

\bibitem{zhang2023deep}
Zhang, Y., Deng, L., Zhu, H., Wang, W., Ren, Z., Zhou, Q., Lu, S., Sun, S., Zhu, Z., Gorriz, J.M., et~al.: Deep learning in food category recognition. Information Fusion  \textbf{98},  101859 (2023). \doi{10.1016/j.inffus.2023.101859}

\bibitem{zhao2023aliked}
Zhao, X., Wu, X., Chen, W., Chen, P.C., Xu, Q., Li, Z.: Aliked: A lighter keypoint and descriptor extraction network via deformable transformation. IEEE Transactions on Instrumentation and Measurement  \textbf{72},  1--16 (2023). \doi{10.1109/TIM.2023.3271000}

\bibitem{zhao2022alikeaccuratelightweightkeypoint}
Zhao, X., Wu, X., Miao, J., Chen, W., Chen, P.C., Li, Z.: Alike: Accurate and lightweight keypoint detection and descriptor extraction. IEEE Transactions on Multimedia  \textbf{25},  3101--3112 (2022). \doi{10.1109/TMM.2022.3155927}

\end{thebibliography}

\end{document}